\documentclass{article} 
\usepackage{times}  
\usepackage{helvet}  
\usepackage{courier}  
\usepackage[hyphens]{url}  
\usepackage{graphicx} 
\usepackage{ijcai16}

\usepackage{hyperref}
\usepackage{caption} 
\DeclareCaptionStyle{ruled}{labelfont=normalfont,labelsep=colon,strut=off} 
\usepackage{amsmath}
\usepackage{amssymb}
\usepackage{booktabs}

\usepackage{standalone}
\usepackage{subcaption}
\usepackage{color, colortbl}
\definecolor{Gray}{gray}{0.8}
\usepackage{placeins}

\usepackage{latexsym}
\usepackage{graphicx}
\usepackage{hyperref}
\usepackage{lipsum,mwe,cuted}
\usepackage{float}
\usepackage{enumerate}
\def\BibTeX{{\rm B\kern-.05em{\sc i\kern-.025em b}\kern-.08em T\kern-.1667em\lower.7ex\hbox{E}\kern-.125emX}}

\usepackage{algorithm, algorithmicx}
\usepackage[noend]{algpseudocode}

\title{Human-vehicle Cooperative Visual Perception for Autonomous Driving under Complex Road and Traffic Scenarios}
\author{
	Yiyue Zhao$^1$,
	Cailin Lei$^1$,
	Yu Shen$^1$, 
	Yuchuan Du$^1$,
	Qijun Chen$^1$ \\
	$^1$Tongji University\\
	\{sc\_smile, 2010762, yshen, ycdu, qjchen\}@tongji.edu.cn
}
\begin{document}



\maketitle






\section*{\centering Abstract}
\textrm{
Human-vehicle cooperative driving has become the critical technology of autonomous driving, which reduces the workload of human drivers. However, the complex and uncertain road environments bring great challenges to the visual perception of cooperative systems. And the perception characteristics of autonomous driving differ from manual driving a lot. To enhance the visual perception capability of human-vehicle cooperative driving, this paper proposed a cooperative visual perception model. 506 images of complex road and traffic scenarios were collected as the data source. Then this paper improved the object detection algorithm of autonomous vehicles. The mean perception accuracy of traffic elements reached 75.52\%. By the image fusion method, the gaze points of human drivers were fused with vehicles' monitoring screens. Results revealed that cooperative visual perception could reflect the riskiest zone and predict the trajectory of conflict objects more precisely. The findings can be applied in improving the visual perception algorithms and providing accurate data for planning and control.
}

\section{Introduction}
%
%
%
%
With the rapid development of autonomous vehicles(AVs), human-vehicle cooperative driving has become the critical technology. On the one hand, it could improve driving safety and efficiency in urgent and risky situations, such as distraction or incorrect operations of drivers\cite{ref1}\cite{ref2}\cite{ref3}. On the other hand, AVs could assist completing accurate vehicle control and reduce the workload of human drivers \cite{ref4}\cite{ref5}\cite{ref6}. 

Existing studies pay more attention to automatic engineering and drivers’ behaviors\cite{ref1}\cite{ref2}. In the field of automatic engineering, there are several traditional solutions to assist manual driving from the vehicle perspective. For example, Gazit set an autonomous-mode steering wheel and sent a signal to drivers to switch to manual mode\cite{ref9}. In addition, Li proposed a switched cooperative driving model from a cyber-physical perspective to minimize the impact of the transitions between automated and manual driving on traffic operations\cite{ref10}. 

In addition, drivers’ behaviors play an important role in traffic safety. An increasing number of researchers utilize visual, auditory, tactile, and other models to stimulate drivers in the automated mode\cite{ref11}\cite{ref12}\cite{ref13}. Spence studied tactile and multi-sensory spatial warning signals on distracted drivers performs better than a simple sensory\cite{ref14}.

However, existing research seldom studies human-vehicle cooperative driving in the visual perception field. Besides, endless changes occur in the complex road and traffic scenarios, such as complex infrastructures, mixed traffic flow, etc. They bring great challenges to the cooperative system, especially the reliability of visual perception\cite{ref7}. Without accurate detection, the system could hardly make proper decisions and control strategies\cite{ref8}. Thus, accurate and efficient visual perception is essential to the safety of human-vehicle cooperative driving.



\subsection{Visual perception approaches to autonomous driving}

Existing autonomous driving visual perception studies mainly include semantic segmentation and object detection of road scenes. The former perceives the scenes from the global perspective, and it doesn't care about the specific kinematic states of each object. But the latter cares more about changes in objects' position and kinematic states within a continuous time series. So object detection provides more details for subsequent decision-making and control\cite{ref15}.

Since PASCAL VOC competitions in 2012, deep learning has become the most powerful technique to extract objects' features from a massive amount of raw data\cite{ref16}\cite{ref17}. One-stage detectors like YOLO and SSD have the advantages of fast detection and generalized features extraction. Two-stage detectors such as RCNN and FPN have higher accuracy through region proposal\cite{ref18}\cite{ref19}\cite{ref20}\cite{ref21}. However, these algorithms cannot adapt to the complex road and traffic scenarios directly. The Adaptability to complex infrastructures and mixed traffic flow needed to be improved. 


\subsection{Cooperative visual perception}

\subsubsection{Experiment on cooperative perception}
The abovementioned visual perception algorithms almost predict objects with the same weight so that they cannot capture the riskiest zone. Thus, there is a large barrier between autonomous driving visual perception systems and human drivers' visual features. 

Combined with visual sensors and perception algorithms, some scientists designed a human-vehicle cooperative navigation system and simulated visual perception in $\textrm{SCANeR}^{\textrm{TM}} $ Studio \cite{ref22}\cite{ref23}.
Through the precise perception by LIDAR and dynamic command of drivers, the ADAS system can adjust the edge and center of object detection. The experiment revealed that cooperative perception assisted proper decision-making and control in the dangerous scenarios. However, the above process still needed dynamic feedback from human drivers. It decreased the efficiency and doesn't reduce the cognitive load of the human agent.



\subsubsection{Image fusion method for cooperative visual perception}


 
Traditional image fusion methods include crop and paste operations by Pillow, Matplotlib, Scikit-image Packages, etc. However, these methods are likely to lose the features of original images and add signal noise to images during the transformation.

In 1987, Gaussian Pyramid was utilized to fuse images. It executed downsample twice to get the minimized resolution images\cite{ref24}. Through editing frequency band, Gaussian Pyramid conducted a large-scale research, pre-calculate, and image proposal. Similarly, Laplacian Pyramid upsampled and smoothed the minimized image's layer. It deviated from the Gaussian Pyramid in the next layer\cite{ref25}.  By applying Image Pyramid, the detailed pixel information and features could be extracted and maintained. It ensured that images were not distorted during the fusion process. 

\bigskip

\begin{figure}[htbp]
\centerline{\includegraphics[width=3.5in]{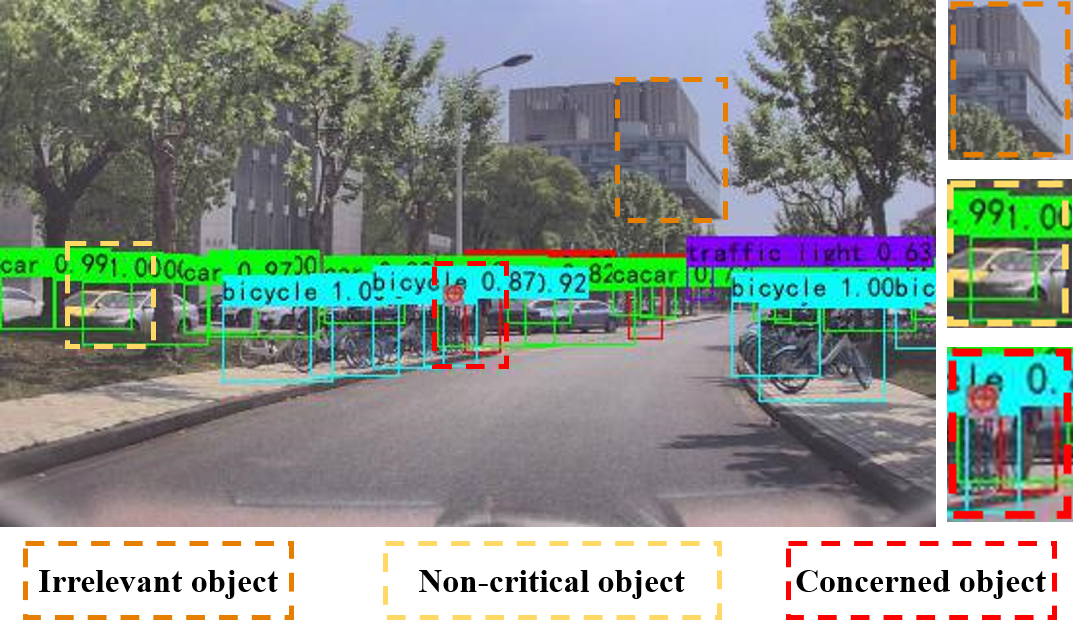}}
\caption{Example of human-vehicle cooperative visual perception. The concerned object is covered by the drivers' gaze points in red.}
\label{fig}
\end{figure}

To bridge the abovementioned research gaps, this paper proposed a visual fusion model for the human-vehicle cooperative perception system, as shown in Figure \ref{fig}. The precision and effect of object detection and image fusion were both verified in the real-world road environment. Based on the transfer learning method and the image fusion algorithm, human-vehicle cooperative visual perception performed better than simple visual perception algorithms. It could predict the trajectory of conflict objects more accurately in complex traffic scenarios, especially the left-turning collision, lane change, and urgent avoidance scenarios\cite{ref26}\cite{ref27}\cite{ref28}.

The rest of this paper is arranged as follows. Section II introduces the critical methodology of human-vehicle cooperative visual perception. The detailed process of the experiment is illustrated in Section IV; Section V concludes this paper with contributions and future studies.


\section{Methodology}

To improve the perception ability and driving safety for AVs in the complex road and traffic scenarios, this study proposes a visual perception fusion method for human-vehicle cooperative driving.

\begin{figure*}

\centering\includegraphics[width=7.0625in]{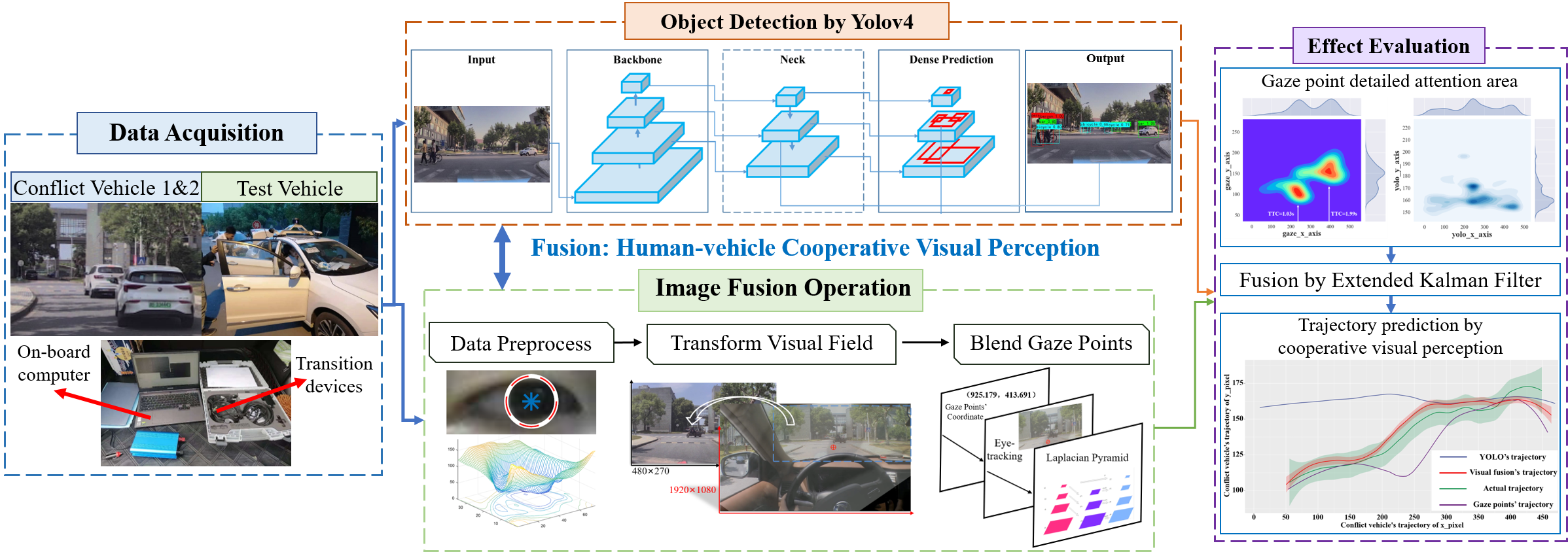}
\captionof{figure}{Framework of Human-vehicle Cooperative Perception.}
\label{method}

\end{figure*}

Cooperative perception includes two steps. First, an object detection algorithm was used to detect the positions and kinematic state of objects under complex road and traffic scenarios. Due to the uncertainty of traffic flow, current object detection algorithms cannot adapt to urgent conflict scenarios. Therefore, this study utilized the transfer learning method to enhance the detection precision in complex road and traffic scenarios. 
Second, this study proposed an image fusion algorithm to fuse drivers' gaze points with in-vehicle camera screens which were predicted by the object detection algorithm. Through analyzing the distribution of drivers' gaze points, the riskiest zone in conflict traffic scenarios could be obtained. The framework of this study is shown in Figure \ref{method}.

\subsection{Data Acquisition}

To obtain the visual characteristics of manual driving in real-world scenarios and vehicles' real-time kinematic states, this study collected multi-source data, including three aspects:

\begin{itemize}
\item Drivers' eye-tracking data

\item Scenario tracking data

\item Real-time kinematic data
 
\end{itemize}

Drivers' eye-tracking data saved the location information of drivers' pupils, which could be abstracted to obtain the gaze points' characteristics. And the scenario tracking data derived from the screen of in-vehicle cameras, which recorded the scenarios in front of the ego vehicle. And the real-time kinematic data returned the high-accuracy information of position and kinematic states, including vehicles' relative displacement, speed, acceleration, etc.

\subsection{Data Preprocessing}
Then, this study processed the collected data. The scenario tracking data and real-time kinematic data would be transformed in the image fusion algorithm. In this part, this paper mainly extracted the drivers' eye-tracking data. By the Canny operator shown as (\ref{i}), all of the boundary points of pupils could be founded easily.

\begin{equation}
\begin{aligned}
	\hat{I} &=\sum_{\omega}a_{guid}I_{guid}+b_{guid} \\
a_{guid} &=\frac{\Phi_{\omega}^{2}}{\Phi_{\omega}^{2}+\varepsilon} \\
b_{guid} &=(1-a_{guid})M_{\omega}
\end{aligned}
\label{i}
\end{equation}

\noindent where

\qquad $I_{guid}$= bootstrap image of the bootstrap filter

\qquad $\hat{I}$= images $I_{guid}$ after filtering

\qquad$a_{guid}$, $b_{guid}$= linearity coefficient

\qquad$\omega$= filter window

\qquad$M_\omega$= the means of $I_{guid}$ in filter window $\omega$

\qquad$\Phi_\omega$= the variance of $I_{guid}$ in filter window $\omega$

\qquad$\varepsilon$= regular term to adjust values of $a_{guid}$

After detecting the boundary of pupils, this study found the location of the pupil center. According to equation $(x-a)^{2}+(y-b)^{2}=r^{2}$, Hough Transform found the circle $(a,b,r)$ that covered most of the boundary point $(x,y)$ in the binary image. Then the pupil center coordinate was determined as $(a,b)$ \cite{ref29}. The Hough Transform algorithm is shown below.

\begin{table}[h]
	\label{table}
	\setlength{\tabcolsep}{3pt}
	\normalsize
	\renewcommand\arraystretch{1.25}
	\begin{tabular}{p{245pt}}
		\hline
		\textbf{Algorithm 1}: Pupil Center Coordinate Determination by Hough Transform\\
		\hline
		Require: \emph{\textbf{L, W, R, x, y, r}}
		\begin{enumerate}[1:]
			\item Determine the sizes of input images: \emph{\textbf{L}}=image.length, \emph{\textbf{W}}=image.width, \emph{\textbf{R}}=$\frac{1}{2} \cdot$ \emph{min}(\emph{\textbf{L,W}}).
			\item Establish a three-dimension array (\emph{\textbf{L,W,R}}) and calculate the number of boundary points (\emph{\textbf{x,y}}) of each parameter circle.
			\item Iterate all boundary points of images, compute circle equations of each boundary point, and record the results into the array: (\emph{\textbf{x,y}}) $\rightarrow$ (\emph{\textbf{L,W,R}})$\rightarrow$ (\emph{\textbf{L,W,R}}).
			\item Find the circle containing the most boundary points (i.e., the Hough circle): $(x-a)^2+(y-b)^2=r^2$.
		\end{enumerate}
	5: Return the pupil center coordinate (\emph{\textbf{L,W,R}}).\\
	    \hline
\end{tabular}
\end{table}

Through \textbf{Algorithm 1}, the location of the pupil’s center was obtained. Taking the geometric center point of two pupils, the trajectory of gaze points could be simplified.

\subsection{Data Fusion Algorithms}

To achieve the human-vehicle cooperative visual perception, this paper established data fusion algorithms between in-vehicle screens after object detection and drivers' gaze points.

\subsubsection{Object Detection of Complex Traffic Scenarios}

The existing object detection algorithms have difficulty perceiving the objects in complex traffic scenarios. The accuracy of object detection would decline dramatically with scenarios’ complexion increasing\cite{ref31}. So this study improved the object detection algorithms based on transfer learning \cite{ref30}.

\begin{figure}[h]
	\centerline{\includegraphics[width=3.5in]{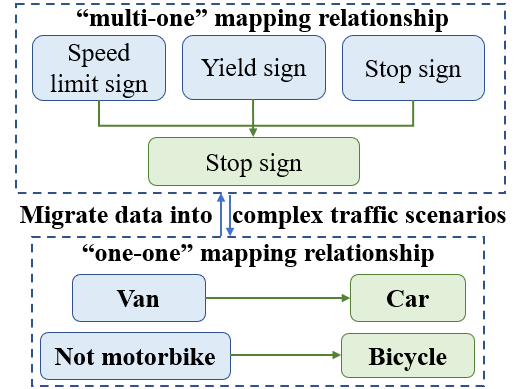}}
	\caption{Data Migration between Coco dataset and complex traffic scenario dataset.}
	\label{migration}
\end{figure}

Transfer learning migrates hyper-parameters from the source domain to the target domain, which effectively utilizes the learning results of the pre-trained model. Thus, the training process requires a smaller volume of datasets and improves efficiency rapidly. The detailed methods could be divided into two steps:


Step 1: migrate data in the source domain to the complex traffic scenario domain. Because the pre-trained model was trained based on the simple samples, it cannot adapt to complex and road traffic scenarios well. Thus, this study established the data migration between the two datasets, including the "many-to-one" mapping relationship and the "one-to-one" derivation relationship, as shown in Figure \ref{migration}.

Step 2: optimize the hyper-parameters of the pre-trained model. Due to the large size of objects in the samples, the pre-trained model hardly recognizes small-size objects in complex road and traffic scenarios. Therefore, this study optimized the hyper-parameters to improve the efficiency and accuracy of detection\cite{ref32}\cite{ref33}. TABLE \ref{tab1} shows the optimized parameters.
\begin{table}[h]
	\caption{Optimized Hyper-parameters and Effect}
	\setlength{\tabcolsep}{3pt}
	\normalsize
	\renewcommand\arraystretch{1.25}
	\begin{tabular}{|p{115pt}|p{125pt}|}
		\hline
		Optimal Hyper-parameter &   Effect\\
		\hline
		Input size:608×608 & Improve boundary accuracy \\
		Regular term: $\alpha=10^{-4}$ & Prevent overfitting\\
		Learning rate:Cosine anneal & Find the optimal solution fast\\
		Mosaic augmentation & Recognize small-size object\\
		label smoothing:0.005 & Reduce classifying uncertainty \\
		Add CIOU to improve loss & Stabilize the target box \\
		\hline
	\end{tabular}
\label{tab1}
\end{table}

\subsubsection{Image fusion Algorithms}

Through data preprocessing, this study obtained the center coordinate $(x_i, y_i)$ of gaze points of human drivers. By measuring the pixel size of the gaze point, the gaze point was represented by a circle that the center coordinate was $(x_i, y_i)$ and radius was 35 pixels. With the Crop function in Pillow (a package of Python), gaze points of human drivers in the eye-tracking videos can be distracted frame by frame. And the crop box would be transformed into coordinate telescoping as shown in (\ref{transform}).

\begin{equation}
	\begin{aligned}
	x_{j}=\frac{(x_i-480)x_1}{x_0-x_1} \pm \frac{35x_1}{2x_0}(\text{min takes -,max takes +}) \\ 
	y_{j}=\frac{(y_i-10)y_1}{y_0-y_1} \pm \frac{35y_1}{2y_0}(\text{min takes -,max takes +}) 
	\end{aligned}
	\label{transform}
\end{equation}

\noindent where

\qquad $x_0$=horizontal axis pixel value of eye-tracking screen

\qquad$x_1$=horizontal axis pixel value of in-vehicle screen

\qquad$y_0$=vertical axis pixel value of eye-tracking screen

\qquad$y_1$=vertical axis pixel value of in-vehicle screen

\qquad$x_i$=horizontal coordinate of the gaze point

\qquad$y_i$=vertical coordinate of the gaze point

\bigskip
Then this study captured the area of eye tracking images arranging from [$x_{jmin}:x_{jmax}, y_{jmin}:y_{jmax}$] and applied Laplacian of Gaussian (LoG) to smooth the signals \textbf{$n_r$}, \textbf{$n_c$} (the length and width of images) from eye-tracking images. The detailed Gaussian Filter\cite{ref24} is shown as (\ref{log}).

\begin{multline}
	LoG(x,y)=\triangledown^{2}G_{\sigma}(x,y)\\
	=\frac{\partial^{2}G_{\sigma}(x,y)}{\partial x^2}+\frac{\partial^{2}G_{\sigma}(x,y)}{\partial y^2}\\
	=-\frac{1}{\pi \sigma^{4}}e^{-\frac{x^2+y^2}{2\sigma^{2}}}(1-\frac{x^2+y^2}{2\sigma^{2}})
	\label{log}
\end{multline}

Based on Algorithm 2 below, this study constructed Gaussian and Laplacian Pyramid\cite{ref25}. Drivers' gaze points of eye-tracking image B were fused with in-vehicle camera screen A after object detection.

\begin{table}[h]
	\setlength{\tabcolsep}{3pt}
	\normalsize
	\renewcommand\arraystretch{1.25}
	\begin{tabular}{p{245pt}}
		\hline
		\textbf{Algorithm 2}: Image fusion Algorithm\\
		\hline
		Require: \boldmath $n_c, n_r, A, B$
		\begin{enumerate}[1:]
			\item Downscale $B$ twice when iterating over the image of the previous layer until the DPI reaches minimization to construct Gaussian Pyramid of $B$: $n_{r}^{'}=\frac{n_r}{2^n}, n_{c}^{'}=\frac{n_c}{2^n}$.
			\item Upscale and smooth the minimized images, deteriorate images of the upper level in the Gaussian Pyramid until the image’s size is equal to the initial image: $LB=B$.laplacian\_pyramid
			\item Repeat the above operation to obtain the Laplace pyramid of $A$: $LA=A$.laplacian\_pyramid
			\item Operate binary mask for $B$ to get interested area (i.e., gaze point): $M=B$.mask
			\item Repeat Step 1 to get the Gaussian Pyramid of $M$: $GM=M$.gaussian\_pyramid
			\item Use $GM$ nodes as weights and synthesize Laplace Pyramid $LS$ from $LA$ and $LB$: $LS(i,j)=GM(i,j)\cdot LA(i,j)+(1-GM(i,j))\cdot LB(i,j)$
		\end{enumerate}
		7: Reconstruct pyramid $LS$ to return fused images $S$: S=reconstruct\_image\_from\_laplacian\_pyramid (LS)\\
		\hline
	\end{tabular}
	\label{tab2}
\end{table}

\subsection{Data Postprocessing and Prediction}

The precise object detection of traffic scenarios and fused images could be further applied to analyze manual eye-tracking characteristics in the process of driving. The object covered by the drivers' gaze point would be given a larger weight. It made the visual perception system better perceive the riskiest zone in the complex traffic scenarios, such as left-turning traffic collision scenarios, lane change scenarios, urgent avoidance scenarios of pedestrians, etc. Through Expand Kalman Filter algorithm, human-vehicle cooperative visual perception could fuse the characteristics of eye-tracking data of human drivers and object detection algorithms.


\section{EXPERIMENT}
\subsection{Setup}

\begin{figure}[h]
	\centerline{\includegraphics[width=3.5in]{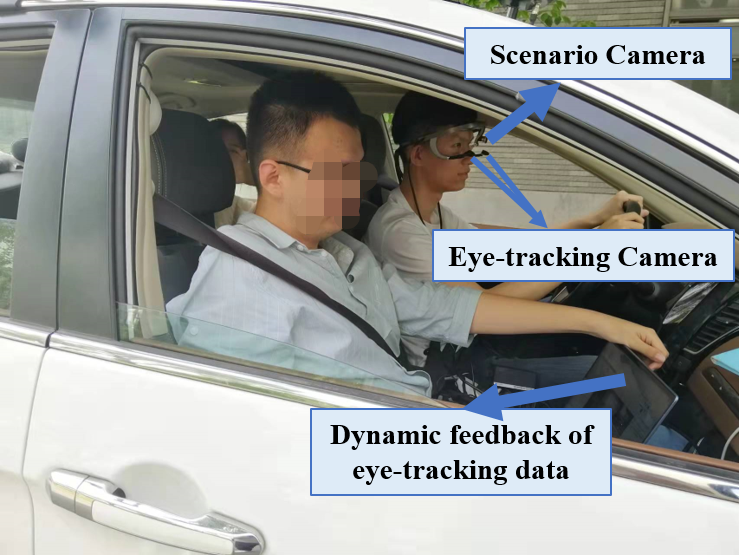}}
	\caption{Eye Tracking Device: Dikablis Professional Glasses.}
	\label{camera}
\end{figure}

The experiment recruited nineteen volunteers with different driving experiences to drive the semi-autonomous vehicle produced by DongFeng. This paper collected data in Tongji University during $8^{th}$ and $9^{th}$ May 2021, covering sunny and cloudy conditions. Participants were required to wear eye tracking device Dikablis Professional Glasses 3.0, as shown in Figure \ref{camera}. It could track pupils’ trajectory during the experiment and return the pupils' position data per 0.117s.

In addition, variables of sensors were installed on the experimental vehicle. In-vehicle camera from Robosense recorded the scenario in front of the experimental vehicle. Real-time kinematic (RTK) from Robosense could return the vehicles' kinematic state information in real-time including vehicles' relative displacement, absolute velocity and acceleration, free space away from the sidewalk, and the distance of deviation from the lane centerline.

To obtain manual eye-tracking features in various complex traffic scenarios, the experiment set up six driving scenarios. They contained three conflict scenarios: left-turning traffic conflict scenarios, lane change scenarios, and urgent avoidance for pedestrians. Table \ref{test} shows the experiment number and description of scenarios.

\begin{table}[h]
	\caption{Experimental Scenario Design}
	\centering
	\setlength{\tabcolsep}{3pt}
	\normalsize
	\renewcommand\arraystretch{1.25}
	\begin{tabular}{|p{80pt}|p{160pt}|}
		\hline
		Scenario's number &  Driving Scenarios \\
		\hline
		\qquad \quad 1 & Straight driving \\
		\qquad \quad 2 & Turn left or right without obstacles\\
		\qquad \quad 3 & Vehicle-vehicle conflict at the cross\\
		\qquad \quad 4 & Two-vehicle or three-vehicle following \\
		\qquad \quad 5 & Conflict between pedestrian and vehicle \\
		\qquad \quad 6 & Conflict between non-motorized vehicles and motorized vehicles \\
		\hline
	\end{tabular}
\label{test}
\end{table}

\begin{figure}[h]
	\centerline{\includegraphics[width=3.5in]{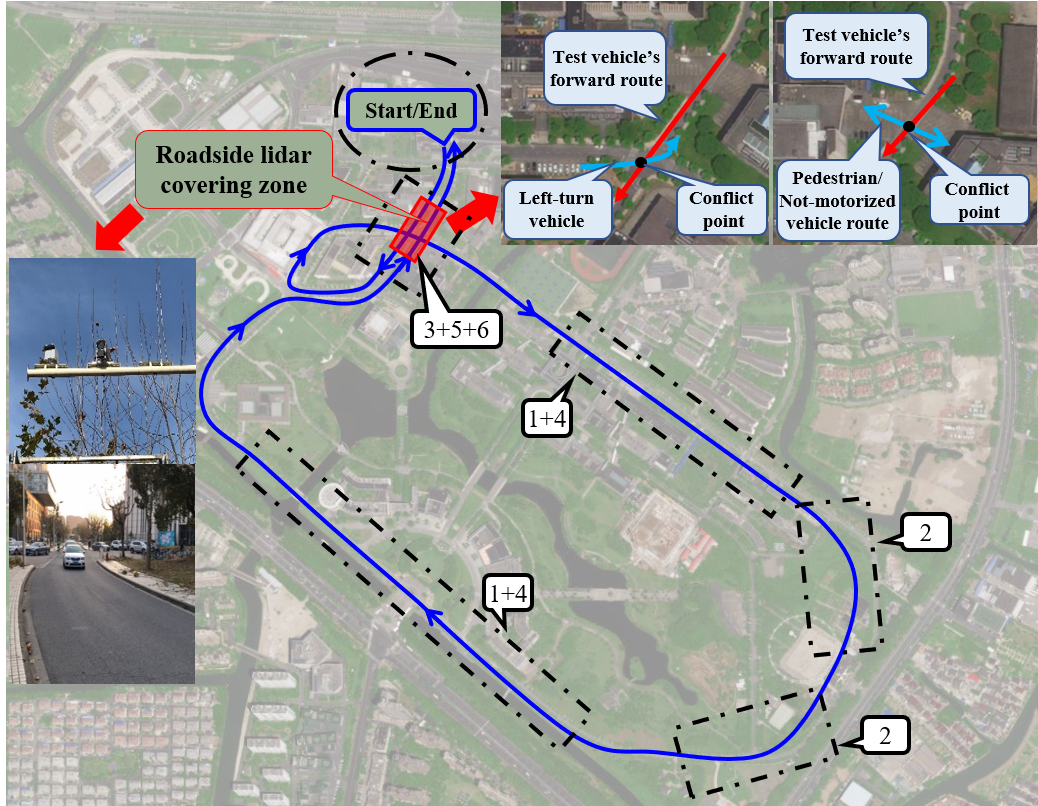}}
	\caption{The illustration of the experimental scenarios.}
	\label{experiment}
\end{figure}

The detailed experimental scenario was located at Tongji University, Shanghai. Based on the scenarios in Table \ref{test}, the experiment constructed a closed route on the campus. Figure \ref{experiment} illustrates the driving route and conflict scenarios.

The nineteen participants included nine males and 10 females(mean ages(M)=21 years, Standard deviation(SD)=1.1 years). The driving experience of participants was arranged from 0.5 to 5 years(M=1.875 years, SD=0.75 years).

Participants were divided into three groups, and there were no obvious demographic differences of participants among these three groups.
The first group needed to tackle the vehicle-vehicle conflict in No.3 experiment. The second group would meet conflicts between pedestrians and vehicles in No.5 experiment. And the third group needed to deal with the conflict between not-motorized vehicles and motorized vehicles in No.6 experiment.


\subsection{Object Detection of Vehicle's Visual Perception}

\subsubsection{Dataset}

Due to the advantages of transfer learning, the required scale of the dataset was small. This paper distracted 505 images of complex traffic scenarios like left-turning traffic collision scenarios, lane change scenarios, and urgent avoidance for pedestrians. Through drawing rectangles, classes of objects were labeled manually.

\subsubsection{Training Detail}

The dataset was divided into train-set and validation-set by 9:1 randomly. The improved backbone of the model was CSPDarkNet53 and the activate function was justified to Mish, as shown in (\ref{mish}).

\begin{equation}
	Mish=x \cdot tanh(ln(1+e^{x}))
	\label{mish}
\end{equation}

The feature enhancement extraction network added Spatial Pyramid Pooling (SPP) and PANet, which could greatly increase the perceptual field, repeatedly extract features, and isolate the most significant contextual features.

\subsubsection{Evaluation Metrics}

The metrics for evaluation included the mean of each of these values: F1-score, Precision Rate, Recall Rate, and Average Precision (AP). Mean Average Precision (mAP) has become a common metric to evaluate the results of multi-classification tasks. It could be calculated by  (\ref{map}).
\begin{equation}
\begin{aligned}
AP&=\frac{1}{11}\sum_{r\in \left\lbrace 0,0.1,\dots,1\right\rbrace }P_{interp}(r)\\
mAP&=\frac{\sum_{q=1}^{Q}AP(q)}{Q}
\end{aligned}
\label{map}
\end{equation}

\subsubsection{Loss Study}

\begin{figure}[h]
	\centerline{\includegraphics[width=3.5in]{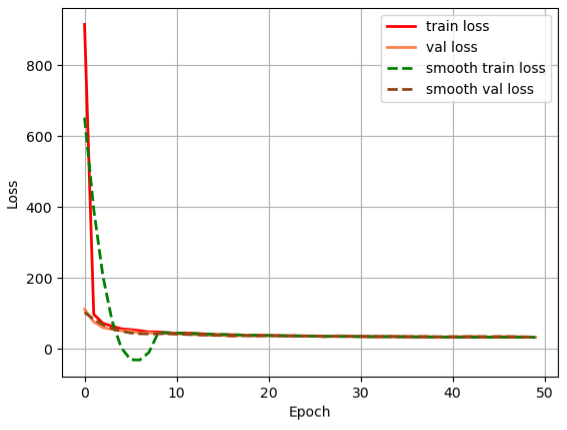}}
	\caption{Loss of the pre-trained model with the weight of the CoCo dataset.}
	\label{loss}
\end{figure}

To assess the validity of the pre-trained model, this paper applied the pre-trained weight of the Coco dataset to train. And the loss result in the process of iterating is shown in Figure \ref{loss}.

As shown in Figure \ref{loss}, the pre-trained model converged fast and the mAP of the Coco dataset was 69.3\%. It ranked in the upper level among the open-source algorithms. 


\subsubsection{Detection Algorithm Improvement by Transfer Learning}

This part would improve the framework of the pre-trained model by using transfer learning, making it better adapt to complex traffic scenarios.

\begin{figure}[h]
	\centering\includegraphics[width=3.5in]{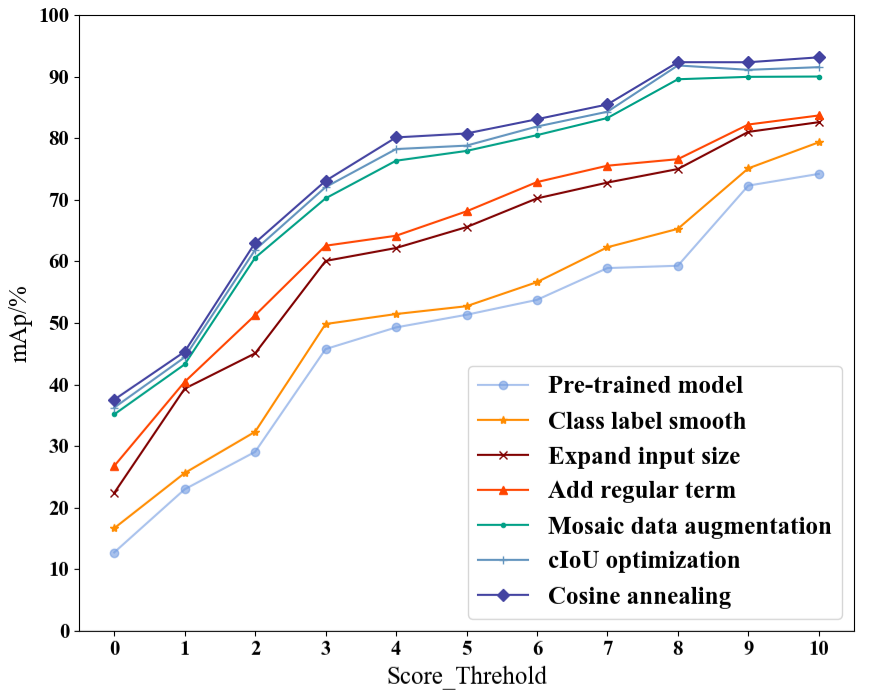}
	\captionof{figure}{Framework improvement by transfer learning.}
	\label{mAP}
\end{figure}

Then, this study utilized the pre-trained model to predict our dataset directly. The mAP of twelve classes remained at 41.7\%, and the problems of missing or wrong predictions were exposed. To improve the accuracy and efficiency of the pre-trained model, this study set control experiments including six experimental groups and one control group. Through adjusting hyper-parameters of the pre-trained model (according to the order in TABLE \ref{tab1}), the predicted mAP of each group was recorded. Figure \ref{mAP} shows the comparison result.

It revealed the effect of transfer learning. The mAP was improved from 41.7\% of the pre-trained model to 75.52\%. Mosaic data augmentation and input size expansion improved the predicted precision most effectively. Then the improved model could distinguish different objects in the complex road traffic scenarios. It could lay a solid foundation for visual fusion in urgent conflict scenarios.

\subsection{Gaze Point Fusion}

Instead of regarding the conflict object as a particle, visual perception of AVs needed to capture the riskiest zone in complex traffic scenarios, i.e., the gathering zone of drivers’ gaze points. So this study fused drivers’ gaze points with in-vehicle camera screens after object detection. The detailed process could be divided into three steps.


\subsubsection{Gaze Point Preprocessing}

In the process of the experiment, eye-tracking device recorded pupils' changes dynamically. And this paper applied Canny Operator to gain the edge information of pupils.

\begin{figure}[h]
	\centering\includegraphics[width=3.5in]{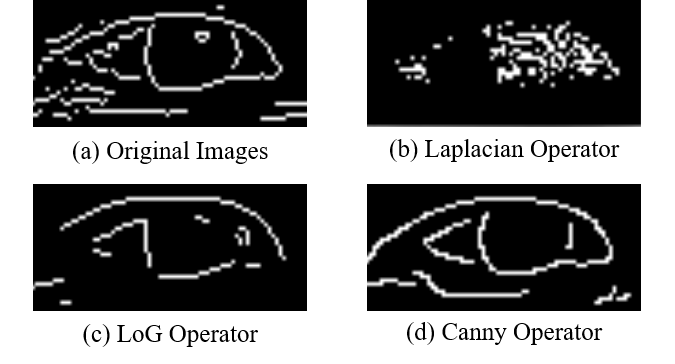}
	\captionof{figure}{Detection comparison of pupils by different  operators.}
	\label{eye}
\end{figure}

Compared with Laplacian Operator and Laplacian of Gaussian Operator(LoG), Canny Operator applies Non-Maximum Suppression to detect the optimal edges of pupils. Through Canny Operator, this study could detect pupils' edges precisely and suppress signal noise. It remained the most comprehensive characteristic of pupils, as seen in Figure \ref{eye}.

After detecting the pupils' edge, this study identified the center of the pupil to obtain precise gaze point further. In this study, the pupil was regarded as a circle. Through Hough Transform and all edge points detected by Canny Operator, the center point and area of pupils could be obtained completely. 


\subsubsection{Coordinate Conversion of Visual Field}

The visual field of the in-vehicle camera screen was different from gaze points, and the eye-tracking device was shaking in the driving process. So this study unified the coordinate between the in-vehicle camera screen and drivers' gaze points.

Through coordinate transforming Equation (\ref{transform}), this study conversed coordinate of gaze points (size=1920$\times$1080) to the coordinate of the in-vehicle screen (size=480 $\times$ 270). Then driver's visual field was identical to the vehicle.

\subsubsection{Gaze Point Extraction and fusion}

Based on the coordinate information of pupils, this study collected the center of gaze points and set 70$\times$ 70 pixels squares to crop gaze points from the eye-tracking screen. Then Gaussian Filter was added to remove signal noise and remain characteristics of gaze points by Laplace Pyramid. Finally, this study applied Equation \ref{transform} to fuse the smoothed gaze points with the in-vehicle camera screen.

\begin{figure}[h]
	\centering\includegraphics[width=3.5in]{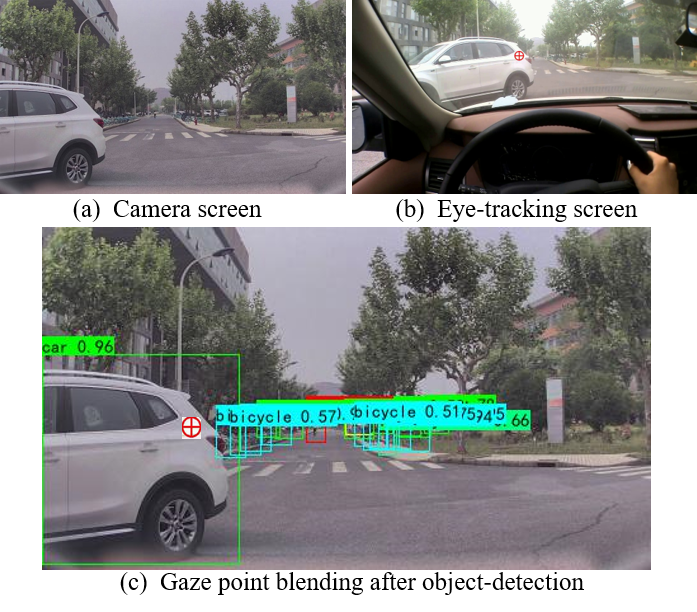}
	\captionof{figure}{Process of visual fusion.}
	\label{fuse}
\end{figure}

Through the three steps, the visual characteristics between human drivers and vehicles were fused. Figure \ref{fuse} illustrates an example of visual fusion in a left-turn conflict scenario. 

\subsection{Ground-truth processing}

To verify the advantages of human-vehicle cooperative visual perception for autonomous driving, this paper utilized Real-time Kinetic(RTK) data to analyze the accuracy of the visual fusion method. Because the results of human-vehicle visual perception are 2D images or videos, this paper transformed relative displacement in the real-world road into pixel-level trajectory to validate in the same coordinate. The transforming formula is shown as Equation \ref{eq:ground}.

\begin{equation}
\begin{aligned}
	p_{x_j}=\frac{s_{x_2}-s_{x_1}}{p_{x_2}-p_{x_1}} \cdot \frac{(x_i-480)x_1}{x_0-x_1} \cdot s_{x_j} \\
	p_{y_j}=\frac{s_{y_2}-s_{y_1}}{p_{y_2}-p_{y_1}} \cdot \frac{(y_i-10)y_1}{y_0-y_1} \cdot s_{y_j}
\end{aligned}
\label{eq:ground}
\end{equation}

\noindent where

\qquad $p_{x_j}$=horizontal value of trajectory at the pixel level

\qquad $p_{y_j}$=vertical value of trajectory at the pixel level

\qquad $s_{x_j}$=horizontal value at the geographic coordinate

\qquad $s_{y_j}$=vertical value at the geographic coordinate

\subsection{Validation of Human-vehicle Cooperative Visual Perception}

Furthermore, this study validated the effect of human-vehicle cooperative visual perception and explored differences in visual characteristics between vehicles' perception systems and manual drivers.

\subsubsection{Experimental Scenario}

\begin{figure}[htbp]
	\centering\includegraphics[width=3.5in]{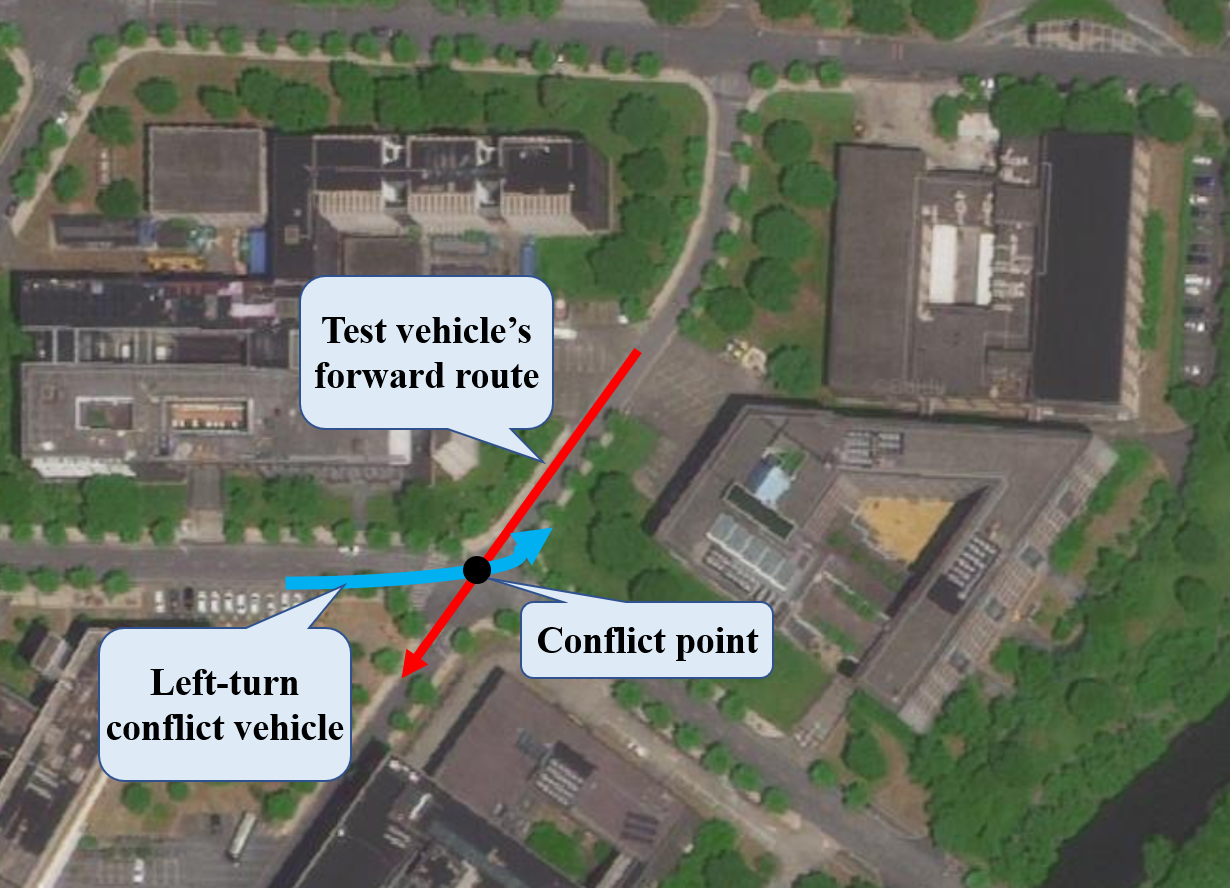}
	\captionof{figure}{Experimental scenario of validation.}
	\label{left-turn}
\end{figure}

This study selected the left-turn conflict scenario to validate the effect of human-vehicle cooperative visual fusion perception. The experimental scenario was shown in Figure \ref{left-turn}. The ego vehicle moved forward while another vehicle turned left suddenly at the intersection.

\begin{figure}[htbp]
	\centering\includegraphics[width=3.5in]{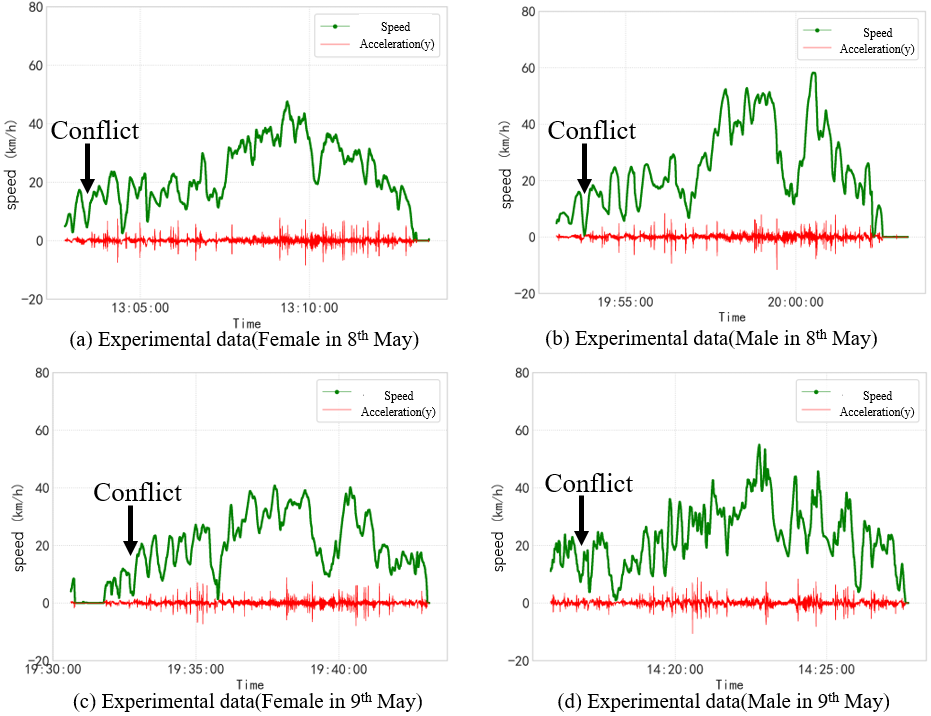}
	\captionof{figure}{Changes of velocity and acceleration in longitudinal direction among four experiments.}
	\label{exp-data}
\end{figure}

And RTK returned the states of the experimental vehicle and the closest conflict object dynamically.
Figure \ref{exp-data} shows the detailed velocity and acceleration change of test vehicles. Additionally, RTK also provided the kinematic states of the closest conflict objects and the distance between ego vehicle and the closest conflict objects.

\subsubsection{Evaluation Metrics}

Time to Collision(TTC) measures the urgent degree in the conflict scenario. First order of TTC could be calculated by Equation \ref{ttc}:

\begin{equation}
TTC=\frac{\Delta S}{\Delta v}
\label{ttc}
\end{equation}

\noindent where:

$\Delta S$= Distance between test vehicles and conflict object

$\Delta v$= Relative Speed derivation in the longitudinal direction 

In the left-turn conflict scenarios, $\Delta v= v_{1y}+v_{2y}$. And $v_{1y}, v_{2y}$ indicated the speed in the longitudinal direction of the ego vehicle and left-turn conflict vehicles. Then this study obtained the gathering zone of gaze points under different TTC conditions.

\subsubsection{Trajectory and Attentive Zone of Human's Gaze Points}

\begin{figure}[h]
	\centering\includegraphics[width=3.5in]{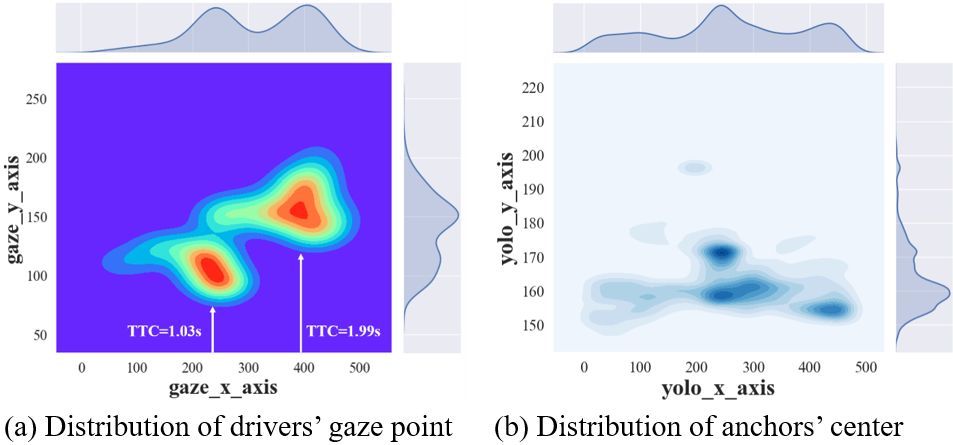}
	\captionof{figure}{Differences between drivers' gaze zone and object detection zone by YOLO.}
	\label{vertify}
\end{figure}

This study collected 36 left-turn conflict scenarios (like Figure \ref{left-turn}) from 19 participants. And the distribution of drivers' gaze points and the center of anchors were obtained. Figure \ref{vertify} shows the comparison between gaze points of human drivers and AVs' visual perception.

\begin{figure*}
\centering\includegraphics[width=7.0625in]{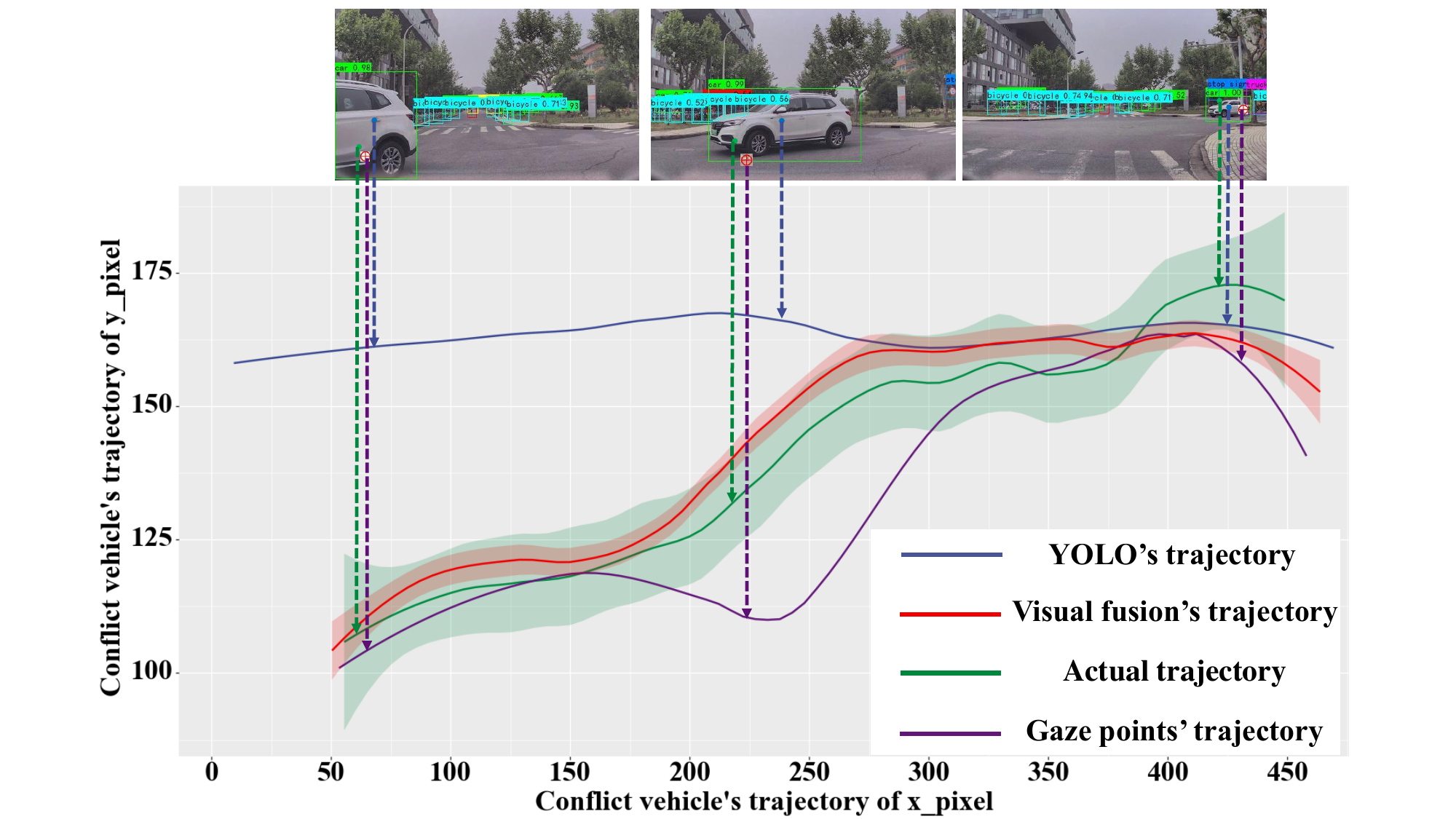}
\captionof{figure}{Actual trajectory and predicted trajectory after visual fusion.}
\label{result}
\end{figure*}

Figure \ref{vertify} suggested that when TTC was beyond 2.0s, drivers' gaze points were nearly out of the closest conflict vehicles. When TTC declined below 2.0s, drivers tended to care farther zone of the conflict car(like the rear tire of vehicles). While TTC $\leq$ 1.03s, drivers tended to care the closet zone away from them, as shown in Figure \ref{vertify}(a). In contrast, the distribution of predicted anchors' centers almost remained at a certain height, without the obvious attentive zones.

Unlike existing object detection algorithms identifying conflict objects' positions by the center point of predicted anchors, human-vehicle coordinate visual perception cares about the detailed contour position of conflict objects. It indicates the riskiest zone in the urgent conflict scenarios. To improve the accuracy of visual perception for AVs, it's essential to fuse gaze points and object detection result.

\subsubsection{Trajectory prediction}

Based on the result above, this paper further assessed the validation of cooperative visual perception by comparing the actual ground-truth trajectory and the trajectory after visual fusion.

\begin{itemize}
\item Actual trajectory

By transforming real-world relative displacement into pixel-level trajectory by Equation \ref{eq:ground}, the prediction effect of trajectory could be evaluated at the same coordinate axis. 

\item Gaze points' trajectory

\item Yolo's trajectory

\end{itemize}

Based on the trajectory of gaze points and YOLO, this paper applied Extended Kalman Filter(EKF) to fuse these two trajectories. The fused trajectory was determined as the predicted trajectory of conflict objects. The Extended Kalman Filter design is shown as Equation \ref{ekf}.

\begin{equation}
	\begin{aligned}
	&\hat{z}_{k+1}= h(\hat{x}_{k+1|k}) \\
	&\hat{x}_{k+1|k+1}= \hat{x}_{k+1}+K_{k+1}(z_{k+1}-\hat{z}_{k+1|k}) \\
	&\hat{x}_{k+1}= f(\hat{x}_{k|k}) \\
	&K_{k+1}=P_{k+1|k}H_{k+1}^T(H_{k+1}P_{k+1|k}H_{k+1}^T+R_{k+1})^{-1} \\
	&P_{k+1|k}=F_kP_{k|k}F_k^T+Q_k \\
	&P_{k+1|k+1}=(I-K_{k+1}H_{k+1})P_{k+1|k}
	\end{aligned}
	\label{ekf}
\end{equation}

\noindent where

\qquad $x_{k|k}$: state estimate value at the moment $k$

\qquad$z_{k+1}$: measured vector value at the moment $k+1$

\qquad$f(x_k)$: non-linear function of the state vector

\qquad$h(x_k)$: non-linear function of measured vector

\qquad$P_{k+1|k}$: state estimate covariance at the moment $k+1$

\qquad$K_{k+1}$: state matrix at the moment $k+1$

\qquad$F_k$: partial derivative of $f(x_k)$ to x at the moment $k$

\qquad$H_k$: partial derivative of $h(x_k)$ to x at the moment $k$

Based on the non-linear function $f(x_k)$ of state vector $x_k$(including gaze point trajectory and YOLO's trajectory) and non-linear function $h(x_k)$ of measured vector $z_k$(fusion trajectory), this study could approximate the next state estimate $z_{k+1}$ of the system through Kalman Filter algorithm. Figure \ref{result} shows the prediction result of the conflict object.

It revealed that the red smooth line of visual fusion's trajectory was closer to the actual trajectory in green than YOLO's trajectory and gaze points' trajectory alone. The Root Mean Squared Error(RMSE) of the predicted trajectory was 0.88 pixel$^2$. 

So human-vehicle cooperative visual perception could better predict the trajectory and kinematic states of conflict objects. The accurate prediction could support the decision-making and control of AVs.

\section{CONCLUSION}
This study proposed a visual fusion perception method for human-vehicle cooperative driving under complex road and traffic scenarios. Based on the transfer learning method, this study improved the object algorithm YOLOv4 and made it better adapt to perceive the critical elements in the complex scenarios. The mAP of critical traffic elements was improved to 75.52\%. Then, this paper proposed a visual fusion perception method between drivers' gaze points and in-vehicle camera screens after object detection. It was 
evaluated on the left-turn conflict experiment. Results suggested that the method could predict the conflict object's trajectory more accurately than the simple detection algorithm. The findings could analyze the potential hazards and provide accurate data for decision-making and control for AVs.




For future research, this study can add a gyroscope to the eye-tracking devices to record the rotation angle of drivers' head. It can obtain more precise position and change of drivers' gaze points and reinforce the automation of visual fusion perception. It is beneficial to the reliability of human-vehicle cooperative visual perception.

\section*{Acknowledgment}

This study is sponsored by the National Key Research and Development Program of China under Grant No. 2020AAA0108101.


\begin{thebibliography}{00}

\bibitem{ref1} R. Ma and D. Kaber, "Situation awareness and workload in driving while using adaptive cruise control and a cell phone," International Journal of Industrial Ergonomics, vol. 35, no. 10, pp. 939-953, 2005.

\bibitem{ref2} J. Li, L. Yao, X. Xu, B. Cheng, and J. Ren, "Deep reinforcement learning for pedestrian collision avoidance and human-machine cooperative driving," Information Sciences, vol. 532, pp. 110-124, 2020.

\bibitem{ref3} J. Stender-Vorwachs and H. Steege, "Legal aspects of autonomous driving," Internationales verkehrswesen, vol. 70, no. MAYSPEC., pp. 18-20, 2018.

\bibitem{ref4} R. Bennett, R. Vijaygopal, and R. Kottasz, "Attitudes towards autonomous vehicles among people with physical disabilities," Transportation Research Part A Policy and Practice, vol. 127, pp. 1-17, 2019.

\bibitem{ref5} G. H. d. A. Correia and B. van Arem, "Solving the User Optimum Privately Owned Automated Vehicles Assignment Problem (UO-POAVAP): A model to explore the impacts of self-driving vehicles on urban mobility," Transportation Research Part B: Methodological, vol. 87, pp. 64-88, 2016.

\bibitem{ref6} M. R. Endsley, "Toward a Theory of Situation Awareness in Dynamic Systems," Human Factors, vol. 37, no. 1, pp. 32-64, 1995.

\bibitem{ref7} T. Huang, R. Fu, and Y. Chen, "Deep Driver Behavior Detection Model Based on Human Brain Consolidated Learning for Shared Autonomy Systems," Measurement, vol. 179, p. 109463, 2021.

\bibitem{ref8} W. Jinqiang, H. Hang, Z. Peng, S. Zebang, and Z. Qingguo, "Review of development and key technologies in automatic driving," Dianzi Jishu Yingyong, vol. 45, no. 6, 2019.

\bibitem{ref9} R. Y. Gazit, "Steering wheels for vehicle control in manual and autonomous driving," U.S.Patent 20130002416 A1, 2013. 

\bibitem{ref10} Y. Li, D. Sun, M. Zhao, D. Chen, S. Cheng, and F. Xie, "Switched cooperative driving model towards human vehicle copiloting situation: A cyberphysical perspective," Journal of Advanced Transportation, vol.20,no.18, 2018.

\bibitem{ref11} C. Gold and K. Bengler, "Taking over control from highly automated vehicles," Human Factors and Ergonomics Society Annual Meeting Proceedings, vol. 8, no. 64, 2014.

\bibitem{ref12} I. Politis, S. Brewster, and F. Pollick, "Language-Based multimodal displays for the handover of control in autonomous cars," in Proceedings of the 7th International Conference on Automotive User Interfaces and Interactive Vehicular Applications, 2015. 

\bibitem{ref13} T. Ziemke, K. E. Schaefer, and M. Endsley, "Situation awareness in human-machine interactive systems," Cognitive Systems Research, vol. 46, no. dec., pp. 1-2, 2017.

\bibitem{ref14} C. Spence, "Tactile and multisensory spatial warning signals for drivers," IEEE Trans Haptics, vol.1, no.2, pp. 121-129, 2008.

\bibitem{ref15} C. B. Murthy, M. F. Hashmi, N. D. Bokde, and Z. W. Geem, "Investigations of Object Detection in Images/Videos Using Various Deep Learning Techniques and Embedded Platforms-A Comprehensive Review," Applied Sciences-Basel, vol.10, no.9, 2020.

\bibitem{ref16} Y. LeCun, Y. Bengio, and G. Hinton, "Deep learning," Nature, vol.521, no.7553, pp.436-444, 2015.

\bibitem{ref17} Hinton, G., E., Salakhutdinov, R. "Reducing the Dimensionality of Data with Neural Networks," Science, 2006. no.313, pp. 504–507.

\bibitem{ref18}J. Redmon, S. Divvala, R. Girshick, and A. Farhadi, "You Only Look Once: Unified, Real-Time Object Detection,". In Proceedings of the Conference on Computer Vision and Pattern Recognition, Las Vegas, NV, USA, 2016; pp. 779–788.

\bibitem{ref19} W. Liu et al., "SSD: Single Shot MultiBox Detector," Cham, 2016: Springer International Publishing, pp. 21-37, 2016.

\bibitem{ref20} R. Girshick, J. Donahue, T. Darrell, and J. Malik, "Rich Feature Hierarchies for Accurate Object Detection and Semantic Segmentation," IEEE Computer Society, 2013. 

\bibitem{ref21} T. Y. Lin, P. Dollar, R. Girshick, K. He, B. Hariharan, and S. Belongie, "Feature Pyramid Networks for Object Detection," IEEE Computer Society, 2017.

\bibitem{ref22} A. Bochkovskiy, C. Y. Wang, and H. Liao, "YOLOv4: Optimal speed and accuracy of object detection," in IEEE Computer Vision and Pattern Recognition, 2020.

\bibitem{ref23} K. Yue and A. C. Victorino, "Human-vehicle cooperative driving using image-based dynamic window approach: System design and simulation," in 2016 IEEE 19th International Conference on Intelligent Transportation Systems (ITSC), 2016. 

\bibitem{ref24} Z. Tao, D. Mu, and S. Ren, "Information hiding (IH) algorithm based on Gaussian Pyramid and GHM (Geronimo Hardin Massopust) multi-wavelet transformation," International Journal of Digital Content Technology \& Its Applications, vol. 5, no. 3, pp. 210-218, 2011.

\bibitem{ref25} P. J. Burt and E. H. Adelson, "The Laplacian Pyramid as a compact image code," Computer Vision, vol.31, no.4, pp.671-679, 1987.


\bibitem{ref26} E. Dabbour and S. Easa, "Proposed collision warning system for right-turning vehicles at two-way stop-controlled rural intersections," Transportation Research Part C: Emerging Technologies, vol. 42, pp. 121-131, 2014.

\bibitem{ref27} N. Kalra and S. M. Paddock, "Driving to safety: How many miles of driving would it take to demonstrate autonomous vehicle reliability?" Transportation Research Part A: Policy and Practice, vol.94, pp.182-193, 2016.

\bibitem{ref28} W. Yang, X. Zhang, Q. Lei, and X. Cheng, "Research on longitudinal active collision avoidance of autonomous emergency braking pedestrian system (AEB-P)," Sensors, vol.19, no.21, 2019.

\bibitem{ref29} D. H. Ballard, "Generalizing the Hough Transform to detect arbitrary shapes," Readings in Computer Vision, pp.714-725, 1987.

\bibitem{ref30} A. Lazaric, "Transfer in reinforcement learning: a framework and a survey." Springer Berlin Heidelberg, 2012.

\bibitem{ref31} SON,T. T., Mita S., Takeuchi A.. "Road Detection using Segmentation by Weighted Aggregation based on Visual Information and a Posteriori Probability of Road Regions". IEEE International Conference on Systems, New York, 2008.

\bibitem{ref32} I. Loshchilov and F. Hutter, "SGDR: Stochastic gradient descent with warm restarts," in ICLR 2017, 2016.
 
\bibitem{ref33} T. Ngogia, Y. Li, D. Jin, et.al, "Real-time sea cucumber detection based on YOLOv4-Tiny and Transfer Learning using data augmentation," In Proceedings of Swarm Intelligence, ICSI 2021, Qingdao, China, 2021. 


\end{thebibliography}
\end{document}